\documentclass{bmvc2k}

\usepackage{url}
\usepackage{graphicx}
\usepackage{amsmath}
\usepackage{amssymb}
\usepackage{multirow}
\usepackage{float}

\newcommand{\bmx}[0]{\begin{bmatrix}}
\newcommand{\emx}[0]{\end{bmatrix}}

\newcommand{\vect}[1]{\mathbf{#1}}

\newcommand{\matr}[1]{\mathbf{#1}}

\newcommand{\va}[0]{\vect{a}}
\newcommand{\vb}[0]{\vect{b}}
\newcommand{\vd}[0]{\vect{d}}
\newcommand{\vc}[0]{\vect{c}}
\newcommand{\vo}[0]{\vect{o}}

\newcommand{\vh}[0]{\vect{h}}
\newcommand{\vi}[0]{\vect{i}}
\newcommand{\vv}[0]{\vect{v}}

\newcommand{\vw}[0]{\vect{w}}
\newcommand{\vp}[0]{\vect{p}}
\newcommand{\vf}[0]{\vect{f}}

\newcommand{\mE}[0]{\matr{E}}
\newcommand{\mW}[0]{\matr{W}}

\newcommand{\mU}[0]{\matr{U}}

\newcommand{\mA}{\matr{A}}

\newcommand{\specialcell}[2][c]{%
  \begin{tabular}[#1]{@{}c@{}}#2\end{tabular}}
  
\DeclareMathOperator*{\argmax}{\arg \max}

\usepackage{subfigure} 

\usepackage{hyperref}

\title{Oracle Performance for Visual Captioning}

\addauthor{Li Yao}{li.yao@umontreal.ca}{1}
\addauthor{Nicolas Ballas}{nicolas.ballas@umontreal.ca}{1}
\addauthor{Kyunghyun Cho}{kyunghyun.cho@nyu.edu}{3}
\addauthor{John R. Smith}{jsmith@us.ibm.com}{2}
\addauthor{Yoshua Bengio}{yoshua.bengio@umontreal.ca}{1}

\addinstitution{
 Universit\'{e} de Montr\'{e}al\\
}
\addinstitution{
 IBM T.J. Watson Research\\
}
\addinstitution{
 New York University\\
}

\runninghead{Li Yao \bmvaEtAl}{Oracle Performance for Visual Captioning}

\begin{document}

\maketitle
\begin{abstract}
The task of associating images and videos with a natural language description
  has attracted a great amount of
  attention recently.
  The state-of-the-art results on some of the standard datasets have been pushed
  into the regime where it has become more and more difficult to make
  significant improvements. Instead of proposing new models, this work
  investigates performances that an oracle can obtain.
  In order to disentangle the contribution from visual model from the
  language model, our oracle assumes that high-quality visual concept extractor is available and focuses
  only on the language part.
  We demonstrate the construction of such oracles on MS-COCO, 
  YouTube2Text and LSMDC (a combination of M-VAD and MPII-MD).
  Surprisingly, despite the simplicity of the model and the training procedure,
  we show that current state-of-the-art models fall short when
  being compared with the learned oracle. 
  Furthermore, it suggests the inability of current models
  in capturing important visual concepts in captioning tasks.
\end{abstract}

\section{Introduction}

With standard datasets publicly available, such as COCO and 
Flickr~\citep{lin2014microsoft, hodosh2013framing, young2014image} in image captioning, and 
YouTube2Text, MVAD and MPI-MD~\citep{Guadarrama_youtube2text,Torabi2015, Rohrbach2015}
in video captioning, the field has been progressing in 
an astonishing speed. For instance, the state-of-the-art results on COCO image captioning has 
been improved rapidly from 0.17 to 0.31 in BLEU~\cite{kiros2014unifying, devlin2015exploring, donahue2014long,vinyals2014show, xu2015show, mao2014deep, KarpathyCVPR14, bengio2015scheduled,qiwu2015}. 
Similarly, the benchmark on YouTube2Text has been 
repeatedly pushed from 0.31 to 0.50 in BLEU score~\cite{rohrbach2013, venugopalan2014translating, yao2015capgenvid, 
  Venugopalan2015iccv, Huijuan2015, rohrbach15gcpr, Haonan2015,Nicollas15}.
While obtaining encouraging results, captioning approaches involve large networks, usually leveraging convolution network for the visual part and recurrent network for the language side. It therefore results model with a certain complexity where the contribution of the different component is not clear.

Instead of proposing better models, the main objective of this work is to develop
a method that offers a deeper insight of the
strength and the weakness of popular visual captioning models.
In particular, we propose a trainable oracle that disentangles the contribution of the visual model
from the language model.
To obtain such oracle, we follow the assumption  that the image and video captioning task 
may be solved with two steps~\cite{rohrbach2013,fang2014captions} . Consider the model $P(\mathbf{w}|\mathbf{v})$
where $\mathbf{v}$ refers to usually high dimensional 
visual inputs, such as 
representations of an image or a video, and $\mathbf{w}$ refers to a caption, usually a sentence of natural 
language description. In order to work well, $P(\vw|\mathbf{v})$ needs to form higher level visual concept,
either explicitly or implicitly, based on $\mathbf{v}$ in the 
first step, denoted as $P(\va|\vv)$, followed by a language model that transforms
visual concept into a legitimate sentence, denoted by $P(\vw|\va)$.
$\va$ referes to \emph{atoms} that are visually perceivable from $\vv$.

The above assumption suggests an alternative way to build an oracle.
In particular, we assume the first step is 
\emph{close to perfect} in the sense that visual concept (or hints) is observed with almost 100\% accuracy.
And then we train 
the best language model conditioned on hints to produce captions.

Using the proposed oracle, we compare the current state-of-the-art models against it, 
  which helps to quantify their capacity of visual modeling, a major weakness,
  apart from the strong language modeling.
In addition, when being applied on different datasets, the oracle offers insight on
  the intrinsic difficulty and blessing of 
  them, a general guideline when designing new algorithms and developing new models.  
Finally, we also relax the assumption to investigate the case where visual concept may 
  not be realistically predicted with 100\% accuracy and demonstrate a quantity-accuracy
  trade-off in solving visual captioning tasks.

\section{Related work}
\paragraph{Visual captioning}
The problem of image captioning has attracted a great amount of attention lately.
Early work focused on constructing linguistic templates or syntactic trees based on 
a set of concept from visual inputs~\cite{kuznetsova2012collective, mitchell2012midge, kulkarni2013babytalk}. 
Another popular approach is based on caption retrieval in the embedding space such as 
\citet{kiros2014unifying, devlin2015exploring}. Most recently, the use 
of language models conditioned on visual inputs have been widely studied in the work of 
\citet{fang2014captions} where a maximum entropy language model is used and 
in \citet{donahue2014long,vinyals2014show, xu2015show, mao2014deep, KarpathyCVPR14}
where recurrent 
neural network based models are built to generate natural language descriptions. 
The work of \citet{devlin2015language} advocates to combine both types of language models.
Furthermore, CIDEr \citep{vedantam2014cider} was proposed 
as an alternative evaluation metric for image captioning and is shown to be 
more advantageous compared with BLEU and METEOR.
To further improve the performance, \citet{bengio2015scheduled} suggests a 
simple sampling algorithm during training, which was one of the winning recipes for 
MSR-COCO Captioning challenge \footnote{\url{http://mscoco.org}}, and \citet{jia2015guiding} 
suggests the use of extra semantic information to guide the language generation process.

Similarly, video captioning has made substantial progress recently.
Early models such as \citet{kojima2002, barbu2012, rohrbach2013} 
tend to focus on constrained domains with limited appearance 
of activities and objects in videos. They also rely heavily on 
hand-crafted video features, followed by a template-based or 
shallow statistical machine translation approaches to produce captions.
Borrowing success from image captioning, recent models 
such as \citet{venugopalan2014translating, donahue2014long, yao2015capgenvid, 
Venugopalan2015iccv, Huijuan2015, rohrbach15gcpr, Haonan2015} and
most recently \citet{Nicollas15} have  
adopted a more general encoder-decoder approach with end-to-end parameter tuning. Videos 
are input into a specific variant of encoding neural networks to form a higher level 
visual summary, followed by a caption decoder by recurrent neural networks. 
Training such type of models are possible with the availability of three relatively 
large scale datasets, one collected from YouTube by \citet{Guadarrama_youtube2text}, the 
other two constructed based on Descriptive Video Service (DVS) on movies by \citet{Torabi2015} and 
\citet{Rohrbach2015}. The latter two have recently been combined as 
the official dataset for Large Scale Movie Description Challenge (LSMDC) 
\footnote{\url{https://goo.gl/2hJ4lw}}.

\paragraph{Capturing higher-level visual concept} The idea of using intermediate visual concept to guide the caption generation has been discussed 
in \citet{qiwu2015} in the context of image captioning and in \citet{rohrbach15gcpr} for 
video captioning. Both work trained classifiers on a predefined set of visual concepts, extracted 
from captions using heuristics from linguistics and natural language processing. Our work resembles both of them 
in the sense that we also extract similar constituents from captions. The purpose of this study, however, is 
different. By assuming perfect classifiers on those visual atoms, we are able to establish 
the performance upper bounds for a particular dataset. Note that a simple bound is suggested by 
\citet{rohrbach15gcpr} where METEOR is measured on all the training captions against a particular 
test caption. The largest score is picked as the upper bound. As a comparison, our approach 
constructs a series of oracles that are trained to generate 
captions given different number of visual hints. Therefore, such bounds are clear indication of models' 
ability of capturing concept within images and videos when performing caption generation, instead of 
the one suggested by \citet{rohrbach15gcpr} that performs caption retrieval. 
\section{Oracle Model}\label{sec_formulation}
The construction of the oracle is inspired by the observation that 
$P(\vw|\vv) = \sum_{\va} P_{\theta}(\vw|\va)P(\va|\vv)$
where $\vw=\{\vw_1, \dots, \vw_t\}$ denotes a caption containing a sequence of words having a 
length $t$. $\mathbf{v}$ denotes the visual inputs such as an image or a video. 
$\mathbf{a}$ denotes visual concepts which we call ``atoms''. We have explicitly factorized 
the captioning model $P(\mathbf{w}|\mathbf{v})$ into two parts, $P(\mathbf{w}|\mathbf{a})$, 
which we call the conditional language model given atoms, and $P(\mathbf{a}|\mathbf{v})$, 
which we call conditional atom model given visual inputs. To establish the oracle, we assume that 
the atom model is given, which amounts to treat $P(\mathbf{a}|\mathbf{v})$ as a Dirac delta 
function that assigns all the probability mass to the observed atom $\mathbf{a}$. 
In other words, $P(\mathbf{w}|\mathbf{v}) = P_{\theta}(\mathbf{w}|\mathbf{a})$.

Therefore, with the fully observed $\mathbf{a}$, the task of image and video captioning reduces to the task of 
language modeling conditioned on atoms. This is arguably a much easier task compared with the direct 
modeling of $P(\mathbf{w}|\mathbf{v})$, therefore a well-trained model could be treated as a performance 
oracle of it. 
Information contained in $\mathbf{a}$ directly influences the difficulty of modeling 
$P_{\theta}(\mathbf{w}|\mathbf{a})$. For instance, if no atoms are available, 
$P_{\theta}(\mathbf{w}|\mathbf{a})$ reduces to unconditional language modeling, 
which could be considered as a lower bound of $P(\mathbf{w}|\mathbf{v})$. By increasing the 
amount of information $\mathbf{a}$ carries, the modeling of $P_{\theta}(\mathbf{w}|\mathbf{a})$ 
becomes more and more straightforward.

\subsection{Oracle Parameterization}\label{sec_parameterization}
Given a set of atoms $\va^{(k)}$ that summarize the visual concept appearing in the visual inputs $\mathbf{v}$, 
this section describes the detailed parameterization of the model $P_{\theta}(\mathbf{w}|\va^{(k)})$ 
with $\theta$ denoting the overall parameters. In particular, we adopt the commonly used encoder-decoder framework 
\citep{cho2014learning} to model this conditional based on the following simple factorization
$P_{\theta}(\vw|\va^{(k)}) = \prod_{t=1}^T P_{\theta}(\vw_t|\vw_{<t},\va^{(k)})$.

Recurrent neural networks (RNNs) are natural 
choices when outputs are identified as sequences. 
We borrow the recent success from a variant of RNNs 
called Long-short term memory networks (LSTMs) first introduced in \citet{hochreiter1997long}, formulated as the following
\begin{align}
    \label{equ_lstm}
    \left[
        \begin{array}{c}
            p(\vw_t\mid \vw_{< t}, \va^{(k)})) \\
            \vh_t \\
            \vc_t
        \end{array}
    \right]
    = \psi(\vh_{t-1}, \vc_{t-1}, \vw_{t-1}, \va^{(k)}),
\end{align}
where $\vh_{t}$ and $\vc_t$ represent the RNN state and memory of LSTMs at 
timestep t respectively.
Combined with the atom representation, Equ. (\ref{equ_lstm}) is implemented as following
\begin{align*}
    \vf_t &= \sigma(\mW_f \mE_w\left[\vw_{t-1}\right] + \mU_f \vh_{t-1} + \mA_f \Phi(\va^{(k)}) + \vb_f), \\
    \vi_t &= \sigma(\mW_i \mE_w\left[\vw_{t-1}\right] + \mU_i \vh_{t-1} + \mA_i \Phi(\va^{(k)}) + \vb_i), \\
    \vo_t &= \sigma(\mW_o \mE_w\left[\vw_{t-1}\right] + \mU_o \vh_{t-1} + \mA_o \Phi(\va^{(k)}) + \vb_o), \\
    \tilde{\vc}_t &= \tanh(\mW_c \mE_w\left[\vw_{t-1}\right] + \mU_c \vh_{t-1} + \mA_c \Phi(\va^{(k)}) + \vb_c), \\
    \vc_{t} &= \vf_t \odot \vc_{t-1} + \vi_t \odot \tilde{\vc}_t, \\
    \vh_t &= \vo_t \odot \vc_t,
\end{align*}
where $\mE_w$ denotes the word embedding matrix, as apposed to the atom embedding matrix $\mE_a$,
$\mW$, $\mU$, $\mA$ and $\vb$ are parameters of the LSTM.
With the LSTM's state $\vh_t$, the probability of the next word in the sequence is
$\vp_t = {\rm softmax}(\mU_p \tanh(\mW_p \vh_t + \vb_p) + \vd)$
with parameters $\mU_p$, $\mW_p$, $\vb_p$ and $\vd$. 
The overall training criterion of the oracle is
\begin{align}\label{equ_cost}
  \theta = \argmax_{\theta}\mathcal{U}_k(\theta) &= \log \prod_{n=1}^N P_{\theta}(\vw^{(n)}|\va^{(n,k)}) =\sum_{n=1}^N \sum_{t=1}^T \log P_{\theta}(\vw^{(n)}_t|\vw^{(n)}_{<t}, \va^{(n,k)}),
\end{align}
given $N$ training pairs $(\vw^{(n)}, \va^{(n,k)})$. $\theta$ represents parameters in 
the LSTM.

\subsection{Atoms Construction}\label{sec_atom}
Each configuration of $\mathbf{a}$ may be associated with
a different distribution $P_{\theta}(\mathbf{w}|\mathbf{a})$, 
therefore a different oracle model. We define configuration 
as an orderless collection of unique atoms. 
That is, $\mathbf{a}^{(k)}=\{a_1, \dots, a_k\}$ where $k$ is the size of the bag and 
all items in the bag are different from each other.
Considering the particular problem of image and video captioning, 
atoms are defined as words in captions that are most related to actions, entities,
and attributes of entities (in Figure \ref{fig_atom_extraction}). 
The reason of using these three particular choices of language components as atoms is not an arbitrary decision.
It is reasonable to consider these three types among the most visually perceivable ones when human 
describes visual content in natural language. We further verify this by conducting a human evaluation procedure 
to identify ``visual'' atoms from this set and show that a dominant majority of them indeed match human visual 
perception, detailed in Section  \ref{sec_human_eval}.
Being able to capture these important concepts is considered as crucial in getting superior 
performance. Therefore, comparing the performance of existing models against this oracle reveals their
ability of capturing atoms from visual inputs when $P(\mathbf{a}|\mathbf{v})$ is unknown.

A set of atoms $\va^{(k)}$ is treated as ``a bag of words''. 
As with the use of word embedding matrix in neural language modeling \citep{bengio2003neural}, 
the $i\mbox{th}$ atom $\va^{(k)}_i$ is used to index the atom embedding matrix 
$\mE_a[\va^{(k)}_i]$ to obtain 
a vector representation of it. Then the representation of the entire set of 
atoms is $\Phi(\va^{(k)})=\sum_{i=1}^k \mE_a[\va^{(k)}_i]$.


\section{Contributing factors of the oracle}
The formulation of Section \ref{sec_formulation} is generic, only relying on the assumption the two-step
visual captioning process, independent of the parameterization in Section \ref{sec_parameterization}.
In practice, however, one needs to take into account several contributing factors to
the oracle.

Firstly, atoms, or visual concepts, may be defined as 1-gram words, 2-gram phrases and so on.
  Arguably a mixture of N-gram representations has the potential to capture more complicated
  correlations among visual concepts. For simplicity,
  this work uses only 1-gram representations, detailed in Section \ref{sec_atom_extraction}.
Secondly, the procedure used to extract atoms needs to be reliable, extracting mainly \emph{visual} concepts,
  leaving out \emph{non-visual} concepts. To ensure this, the procedure used in this work is verified with
  human evaluation, detailed in \ref{sec_human_eval}.
Thirdly, the modeling capacity of the conditional language $P_{\theta}(\vw|\va^{(k)})$ has a direct influence
  on the obtained oracle. Section \ref{sec_parameterization} has shown one example of many possible
  parameterizations.
Lastly, the oracle may be sensitive to the training procedure and its hyper-parameters
(see Section \ref{sec_train}).

While it is therefore important to keep in mind that proposed oracle \emph{conditions} on
the above factors, quite surprisingly, however, with the simplest procedure and parameterization
we show in the experimental section that oracle serves their purpose
reasonably well.
\section{Experiments}
We demonstrate the procedure of learning the oracle on three standard 
visual captioning datasets. 
MS COCO \citep{lin2014microsoft} is the most commonly used benchmark dataset in image captioning.
It consists of 82,783 training and 40,504 validation images.
each image accompanied by 
5 captions, all in one sentence. We follow the 
split used in \citet{xu2015show} where a subset of 5,000 images are used as validation,
and another subset of 5,000 images are used for testing. 
YouTube2Text is the most commonly used benchmark dataset in video captioning. 
It consists of 1,970 video clips, each accompanied with multiple captions. 
Overall, there are 80,000 video and caption pairs. Following \citet{yao2015capgenvid}, it is 
split into 1,200 clips for training, 100 for validation and 670 for testing.
Another two video captioning datasets have been recently introduced in \citet{Torabi2015} and 
\citet{Rohrbach2015}. Compared with 
YouTube2Text, they are both much larger in the number of video clips,
most of which are associated with one or two captions. 
Recently they are merge together for 
Large Scale Movie Description Challenge (LSMDC). 
\footnote{\url{https://goo.gl/2hJ4lw}}
We therefore name this 
particular dataset LSMDC. The official splits contain 91,908 clips for training,  
6,542 for validation and 10,053 for testing.
\subsection{Atom extraction}\label{sec_atom_extraction}
\begin{figure}[!h]
  \begin{minipage}{0.6\textwidth}
    \includegraphics[scale=0.21]{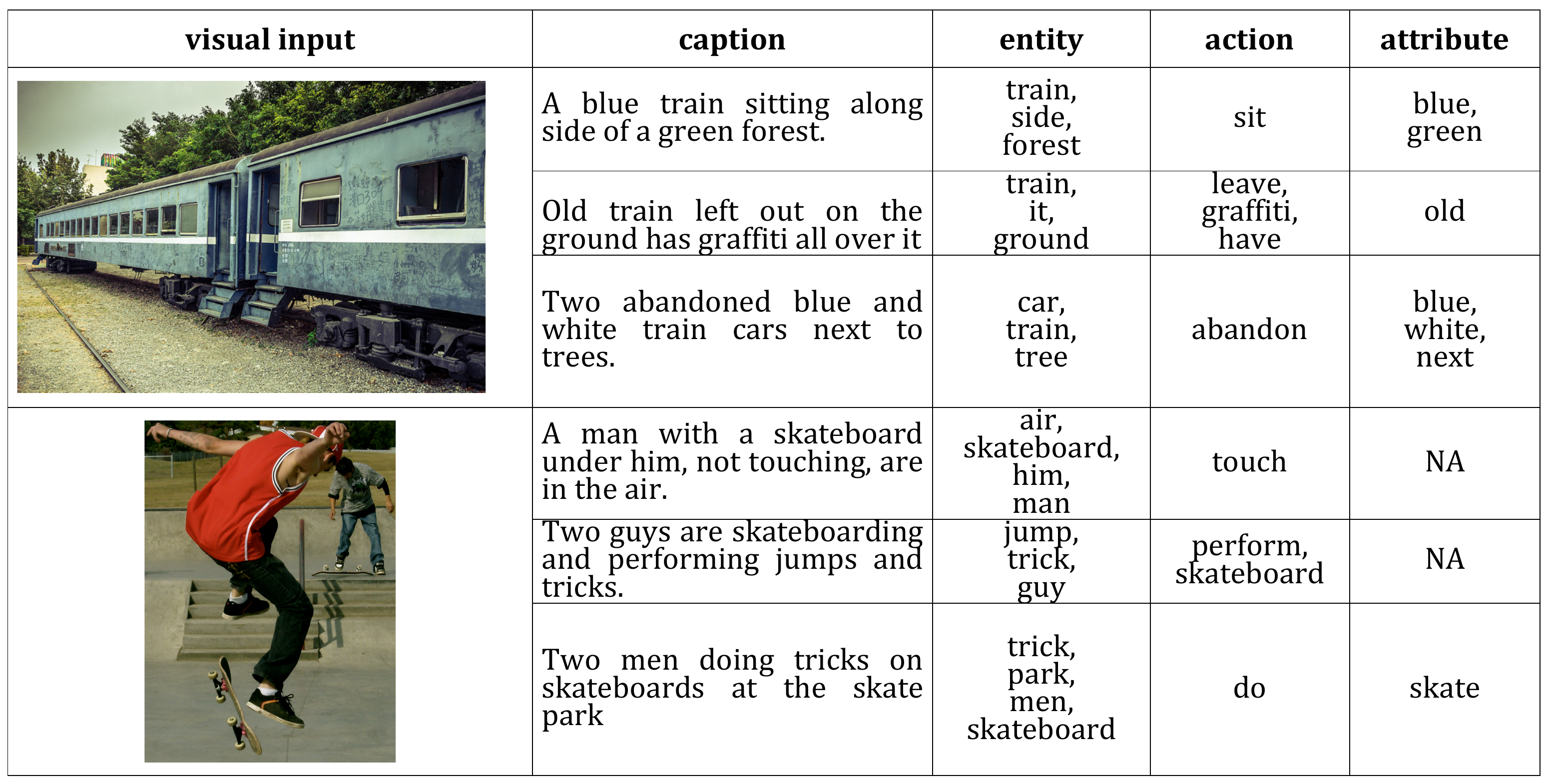}
  \end{minipage}\hfill
  \begin{minipage}{0.4\textwidth}
    \scriptsize{
  \caption{Given ground truth captions, three categories of visual atoms
    (entity, action and attribute) are automatically
    extracted using NLP Parser. ``NA'' denotes the empty atom set.}\label{fig_atom_extraction}}
  \end{minipage}
\end{figure}
Visual concepts in images or videos are summarized as atoms that are provided to the caption language model. 
They are split into three categories: actions, 
entities, and attributes. To identify these three classes, we utilize
Stanford natural language parser 
\footnote{\url{http://goo.gl/lSvPr}} to automatically 
extract them. After a caption is parsed, we apply simple heuristics based on the tags
produced by the parser, ignoring the phrase and sentence level tags
\footnote{complete list of tags: \url{https://goo.gl/fU8zDd}}:
Use words tagged with \{``NN'', ``NNP'', ``NNPS'' ,``NNS'', ``PRP''\} as entity atoms.
Use words tagged with  \{``VB'', ``VBD'', ``VBG'', ``VBN'', ``VBP'', ``VBZ''\} as action atoms.
Use words tagged with  \{``JJ'', ``JJR'', ``JJS''\} as attribute atoms.
After atoms are identified, they are lemmatized with NLTK lemmatizer 
\footnote{\url{http://www.nltk.org/}} to unify them to their 
original dictionary format \footnote{available at 
  \url{https://goo.gl/t7vtFj}}. Figure \ref{fig_atom_extraction} illustrates some results.
We extracted atoms for COCO, YouTube2Text and LSMDC.
This gives 14,207 entities, 4,736 actions and 8,671 attributes for COCO, 
6,922 entities, 2,561 actions, 2,637 attributes for YouTube2Text, and 
12,895 entities, 4,258 actions, 8550 attributes for LSMDC. 
Note that although the total number of atoms of each categories may be large, 
atom frequency varies. In addition, the language parser does not guarantee 
the perfect tags. Therefore, when atoms are being used in training the 
oracle, we sort them according to their frequency and make sure to 
use more frequent ones first to also give priority to atoms with larger coverage, 
detailed in Section \ref{sec_train} below.

\label{sec_human_eval}
We conducted a simple human evaluation
\footnote{details available at \url{https://goo.gl/t7vtFj}}
to confirm that 
extracted atoms are indeed predominantly visual. As it might be impractical 
to evaluate all the extracted atoms for all three datasets, we focus on top 150 
frequent atoms. This evaluation intends to match 
the last column of Table \ref{tab_benchmark} where current state-of-the-art models have the equivalent capacity of 
capturing perfectly less than 100 atoms from each of three categories. Subjects are asked to 
cast their vote independently. The final decision of an atom being visual or not is made by majority vote. 
Table \ref{tab_human_eval} shows the ratio of atoms flagged as visual by such procedure.
\begin{table}[ht]
  \centering
  \small{
\caption{Human evaluation of proportion of atoms that are voted as visual. It is clear that 
extracted atoms from three categories contain dominant amount of visual elements, hence 
verifying the procedure described in Section \ref{sec_atom}. Another observation is that 
entities and actions tend to be more visual than attributes according to human perception.}
\label{tab_human_eval}}
\begin{tabular}{|l||lll|}
\hline
             & entities & actions & attributes \\
\hline
COCO         & 92\%         & 85\%        & 81\%           \\
YouTube2Text & 95\%         & 91\%        & 72\%           \\
LSMDC        & 83\%         & 87\%        & 78\%        \\
\hline  
\end{tabular}
\end{table}

\subsection{Training}\label{sec_train}
After the atoms are extracted, they are sorted according to the frequency they appear 
in the dataset, with the most frequent one leading the sorted list. 
Taking first $k$ items from this list gives the top $k$ most frequent ones, forming 
a bag of atoms denoted by $\va^{(k)}$ where $k$ is the size of the bag. Conditioned 
on the atom bag, the oracle 
is maximized as in Equ (\ref{equ_cost}). 

To form captions, we used a vocabulary of size 20k, 13k and 25k for COCO, {\small YouTube2Text}
and LSMDC respectively. For all three datasets, models were trained on training set 
with different configuration of (1) atom embedding size, (2) word embedding size and 
(3) LSTM state and memory size. To avoid overfitting we also experimented weight 
decay and Dropout \citep{hinton2012improving} to regularize the models with different size. In particular, 
we experimented with random hyper-parameter search by \citet{bergstra2012random} 
with range $[128, 1024]$ 
on (1), (2) and (3). Similarly we performed random search on the weight decay 
coefficient with a range of $[10^{-6}, 10^{-2}]$, and whether or not to use dropout. 
Optimization was performed by SGD, minibatch size 128, and 
with Adadelta \citep{Zeiler-2012} to automatically adjust the 
per-parameter learning rate. Model selection was done on the standard validation set, 
with an early stopping patience of 2,000 (early stop if no improvement made after 2,000 
minibatch updates). We report the results on the test splits.
\subsection{Interpretation}\label{sec_interpretation}
\begin{figure}[t]
\begin{minipage}{0.31\textwidth}
\includegraphics[scale=0.15]{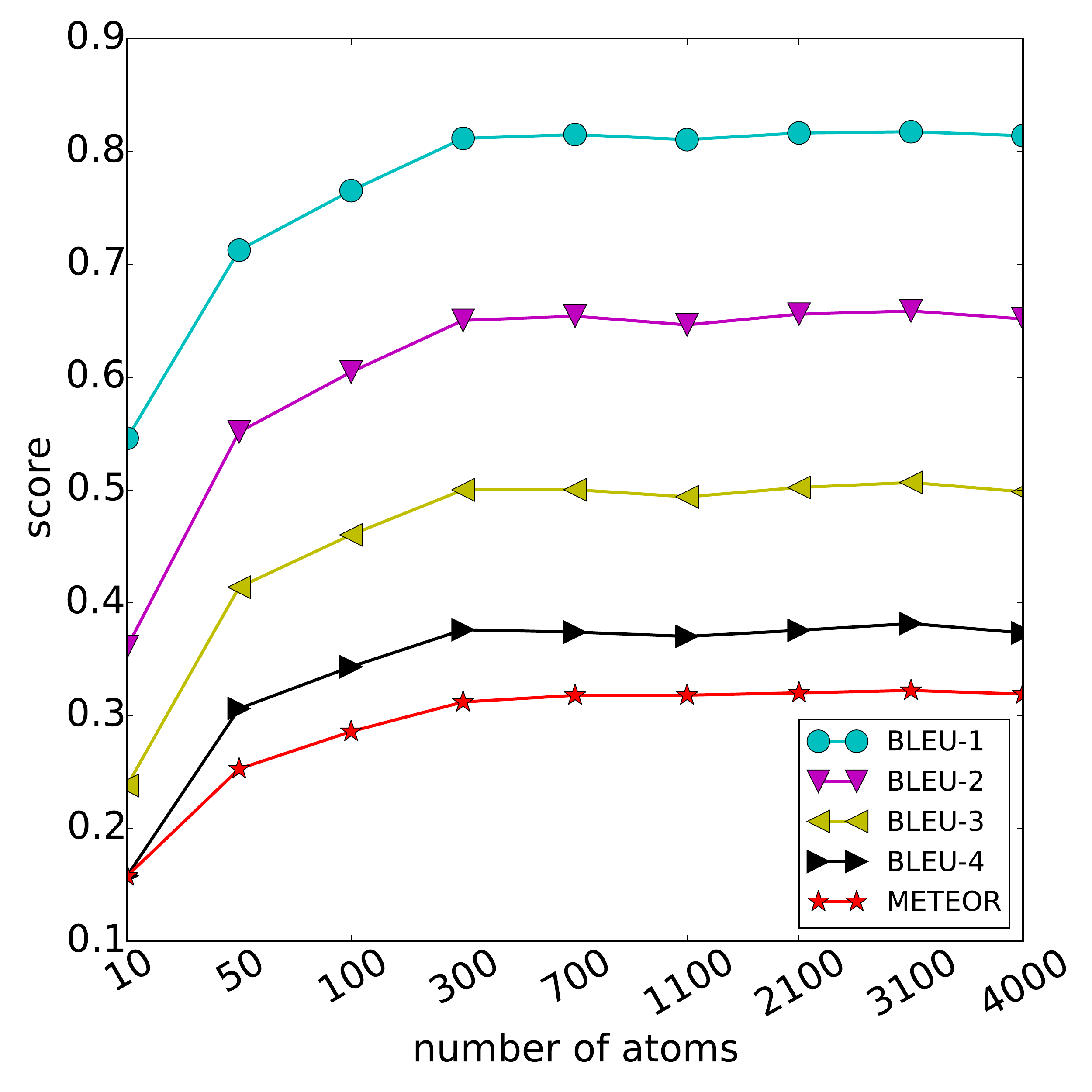}
\end{minipage}
\begin{minipage}{0.31\textwidth}
\includegraphics[scale=0.15]{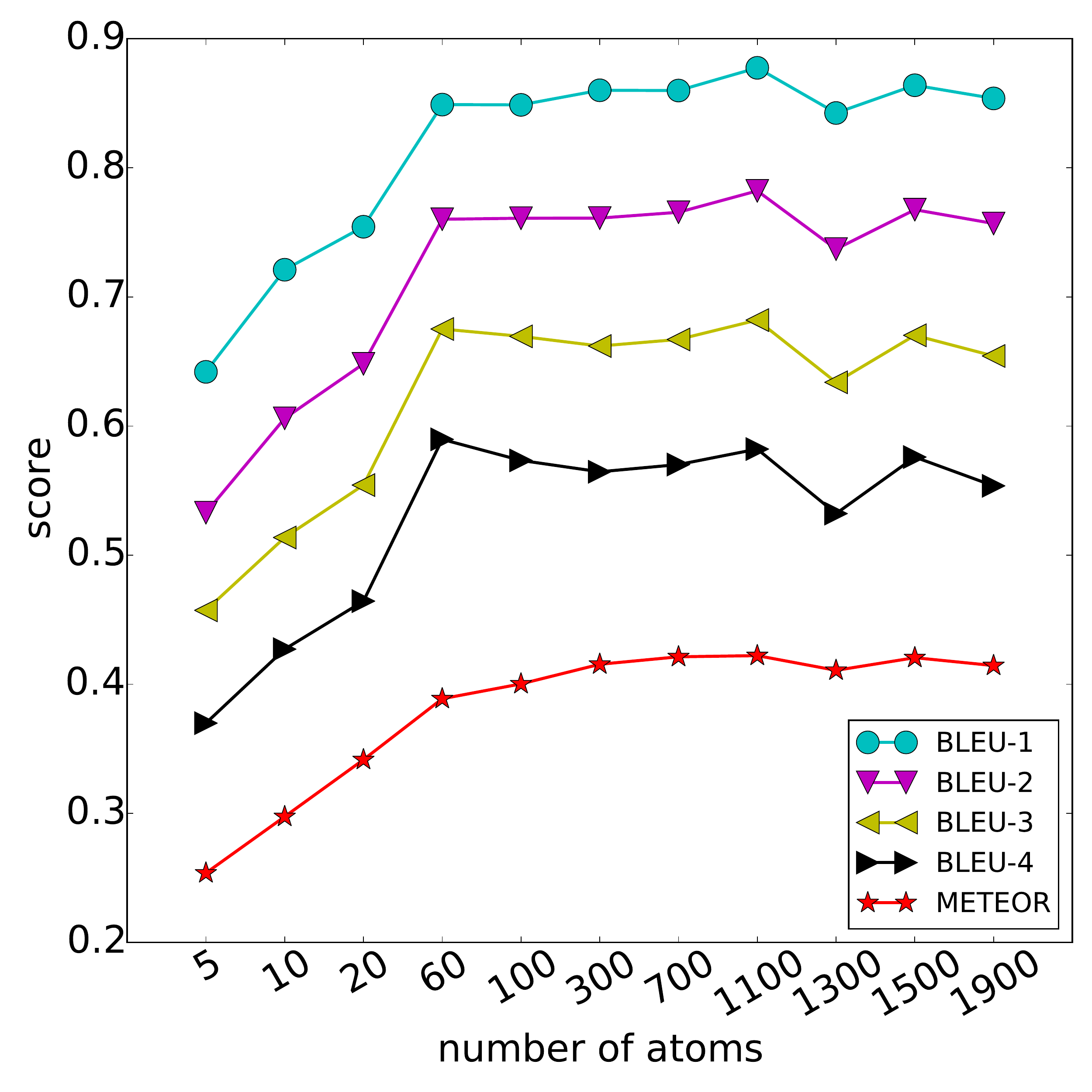}
\end{minipage}\hfill
\begin{minipage}{0.31\textwidth}
  \includegraphics[scale=0.15]{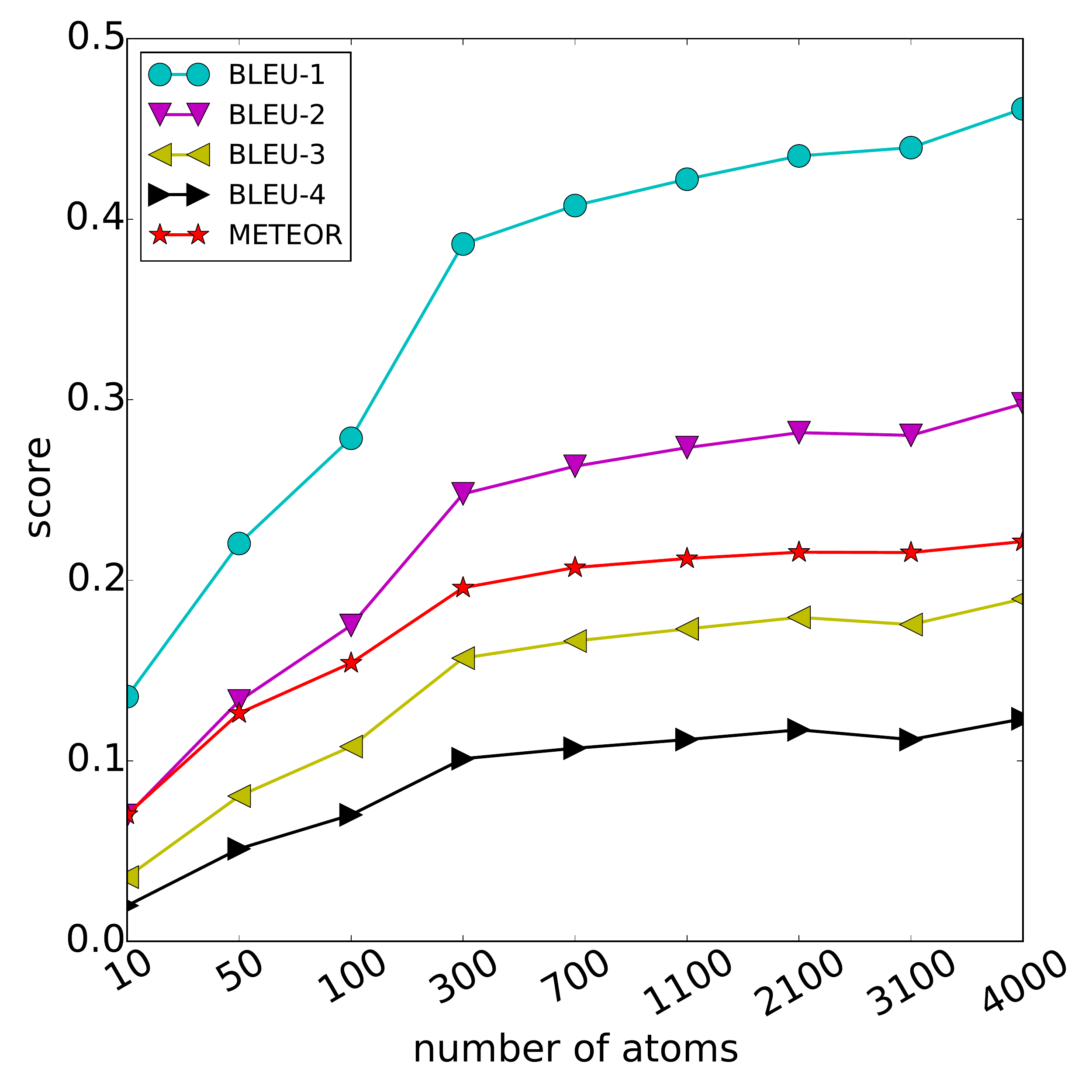}
\end{minipage}
\begin{minipage}{0.31\textwidth}
\includegraphics[scale=0.15]{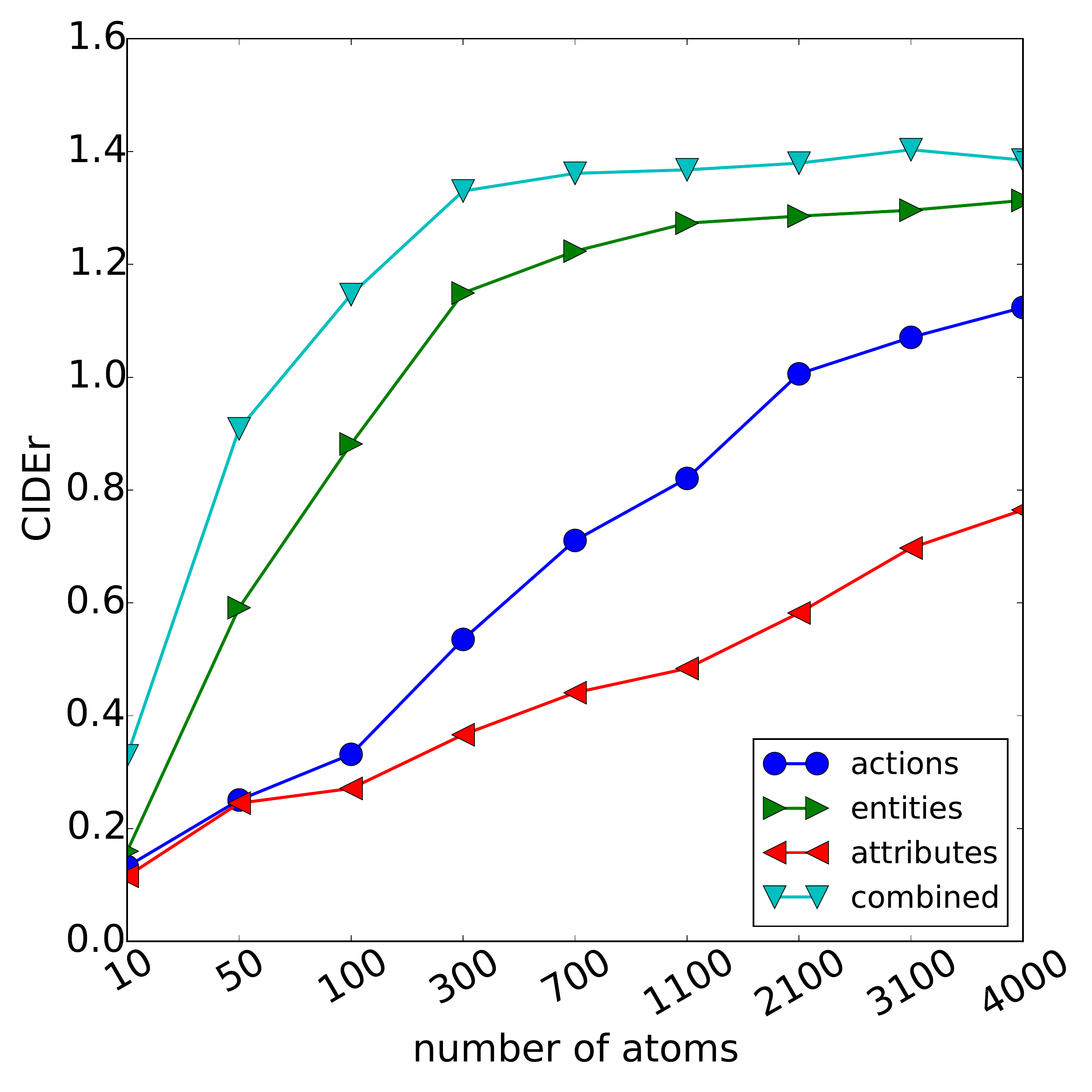}
\end{minipage}
\begin{minipage}{0.31\textwidth}
\includegraphics[scale=0.15]{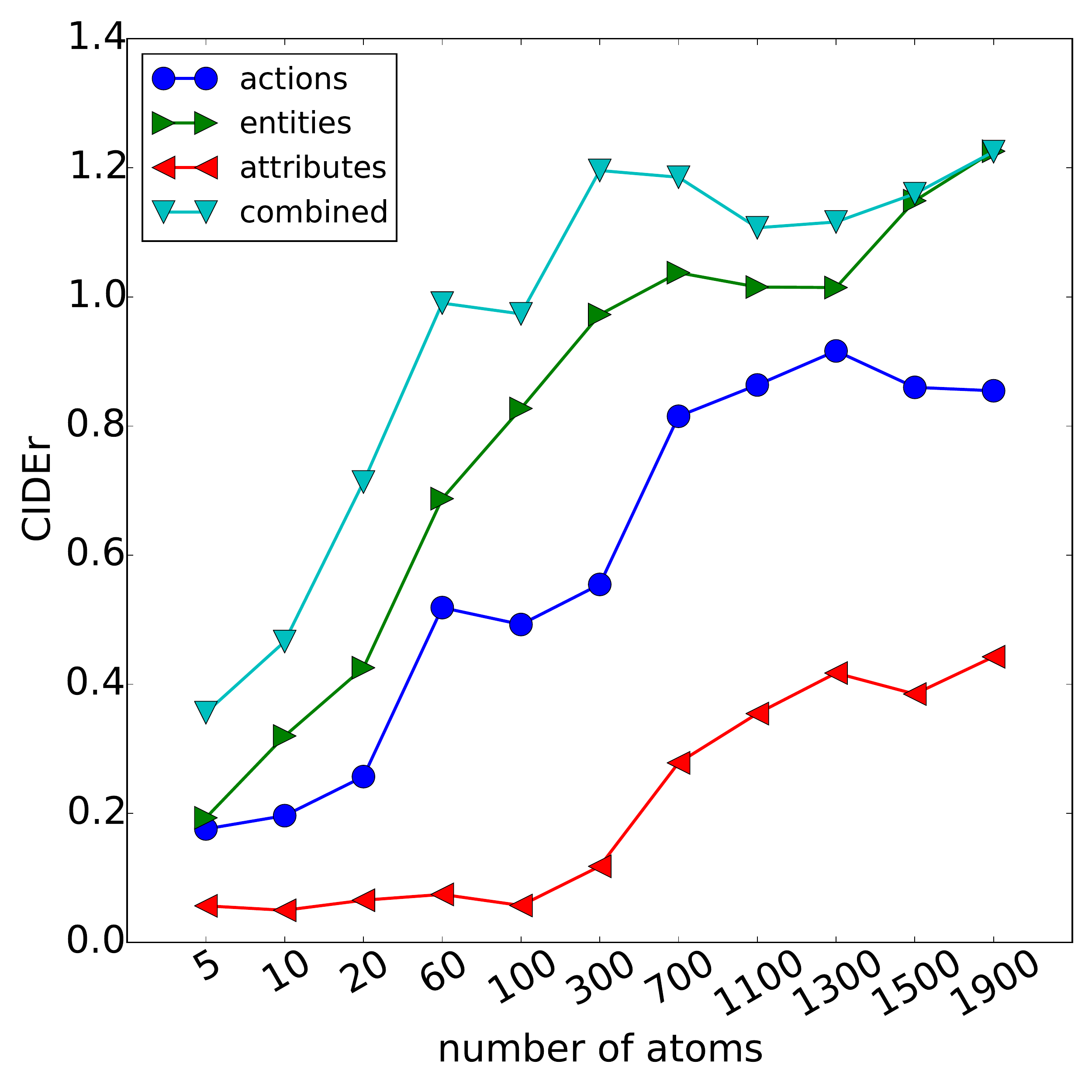}
\end{minipage}\hfill
\begin{minipage}{0.31\textwidth}
\includegraphics[scale=0.15]{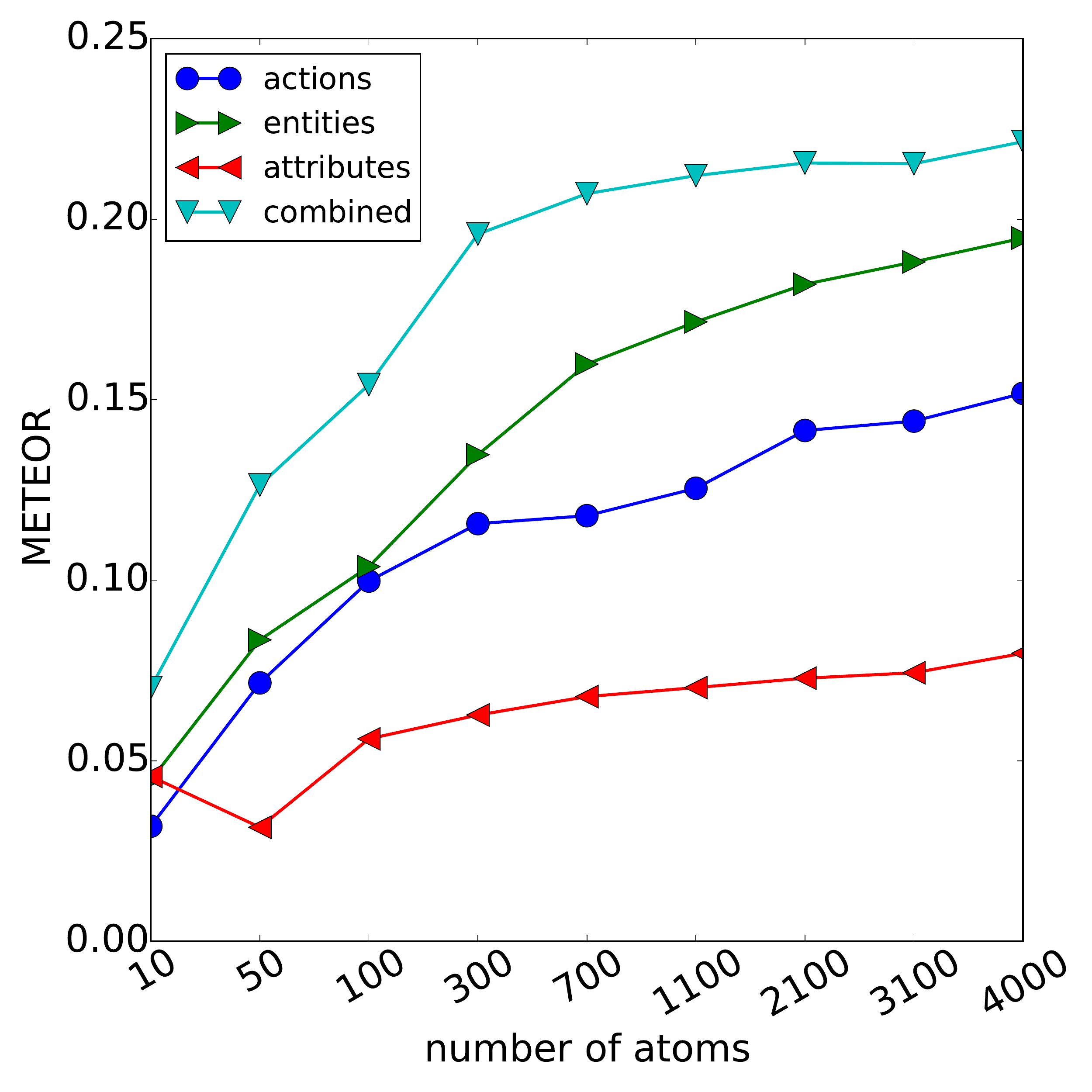}
\end{minipage}
\scriptsize{
\caption{Learned oracle on COCO (left), YouTube2Text (middle) and LSMDC (right).
The number of atoms $\va^{(k)}$ is varied on x-axis and oracles 
are computed on y-axis on testsets. The first row shows the oracles on BLEU and METEOR
with $3k$ atoms, $k$ from each of the three categories. The second row shows the oracles 
when $k$ atoms are selected individually for each category. CIDEr is used for 
COCO and YouTube2Text as each test example is associated with multiple ground truth captions,
argued in \citep{vedantam2014cider}. 
For LSMDC, METEOR is used, as argued by \citet{rohrbach15gcpr}.}
\label{fig_overall}}
\end{figure}

All three metrics -- BLEU, METEOR and CIDER are computed with 
Microsoft COCO Evaluation Server \citep{chen2015microsoft}. Figure \ref{fig_overall} 
summarizes the learned oracle with an increasing 
number of $k$.
\paragraph{comparing oracle performance with existing models}\label{sec_bounding_sota}
We compare the current state-of-the-art models' performance against the 
established oracles in Figure \ref{fig_overall}. Table \ref{tab_benchmark} shows the comparison 
on three different datasets. 
With Figure \ref{fig_overall}, 
one could easily associate a particular performance with the equivalent number of atoms 
perfectly captured across all 3 atom categories, as illustrated in Table \ref{tab_benchmark},
the oracle included in bold. It is somehow surprising that state-of-the-art models have performance 
that is equivalent to capturing only a small amount of ``ENT'' and ``ALL''. This experiment highlights the shortcoming of
the state-of-art visual models. By improving them, we could close the performance gap
that we currently have with the oracles.
\begin{table*}[t]
  \centering
  \scriptsize{
\caption{Measure semantic capacity of current state-of-the-art models.
Using Figure \ref{fig_overall}, one could easily map the reported metric to the number 
of visual atoms captured. This establishes an equivalence between a model, 
the proposed oracle and a model's semantic capacity. (``ENT'' for entities. 
``ACT'' for actions. ``ATT'' for attributes. ``ALL'' for all three categories combined. 
``B1'' for BLEU-1, ``B4'' for BLEU-4. ``M'' for METEOR. ``C'' for CIDEr. 
Note that the CIDEr is between 0 and 10 according to \citet{vedantam2014cider}.
The learned oracle is denoted in \textbf{bold}.}
\label{tab_benchmark}}
\footnotesize
\begin{tabular}{|c||c|c|c|c||c|c|c|c|}
\hline
             & B1       & B4    & M     & C     & ENT & ACT & ATT       & ALL \\
\hline
\specialcell{COCO \\ \citep{qiwu2015}} & \specialcell{0.74\\ \textbf{0.80}}    & \specialcell{0.31\\ \textbf{0.35}}  & \specialcell{0.26\\ \textbf{0.30}}  & \specialcell{0.94\\ \textbf{1.4}}  & $\sim$200     & $\sim$2100   & $>$4000 & $\sim$ 50      \\
\hline
\specialcell{YouTube2Text \\ \citep{Haonan2015}} & \specialcell{0.815\\ \textbf{0.88}}  & \specialcell{0.499\\ \textbf{0.58}} & \specialcell{0.326\\ \textbf{0.40}} &\specialcell{0.658\\ \textbf{1.2}} & $\sim$60      & $\sim$500    & $>$1900 & $\sim$ 20    \\
\hline
\specialcell{LSMDC \\ \citep{Venugopalan2015iccv}} & \specialcell{N/A \\ \textbf{0.45}}    & \specialcell{N/A\\ \textbf{0.12}}   & \specialcell{0.07\\ \textbf{0.22}}  & \specialcell{N/A\\ \textbf{N/A}}   & $\sim$40      & $\sim$50     & $\sim$4000  & $\sim$10 \\
\hline     
\end{tabular}
\end{table*}
\paragraph{quantify the diminishing return}
As the number of atoms $k$ in $\va^{(k)}$ increases, one would expect the oracle 
to be improved accordingly.  
It is however not yet clear the speed of such improvement. 
In other words, the gain in performance may not be proportional to the number of 
atoms given when generating captions, due to atom frequencies and language modeling.
Figure \ref{fig_overall} quantifies this effect. 
The oracle on all three datasets shows a significant gain 
at the beginning and diminishes quickly as more and more atoms are used. 

Row 2 of Figure \ref{fig_overall} also highlights the difference among 
actions, entities and attributes in generating captions. For all three 
datasets tested, entities played much more important roles, even more so than 
action atoms in general. This is particularly true on LSMDC where the gain 
of modeling attributes is much less than the other two categories. 

\begin{figure}[h]
\begin{minipage}{0.3\textwidth}
\includegraphics[scale=0.15]{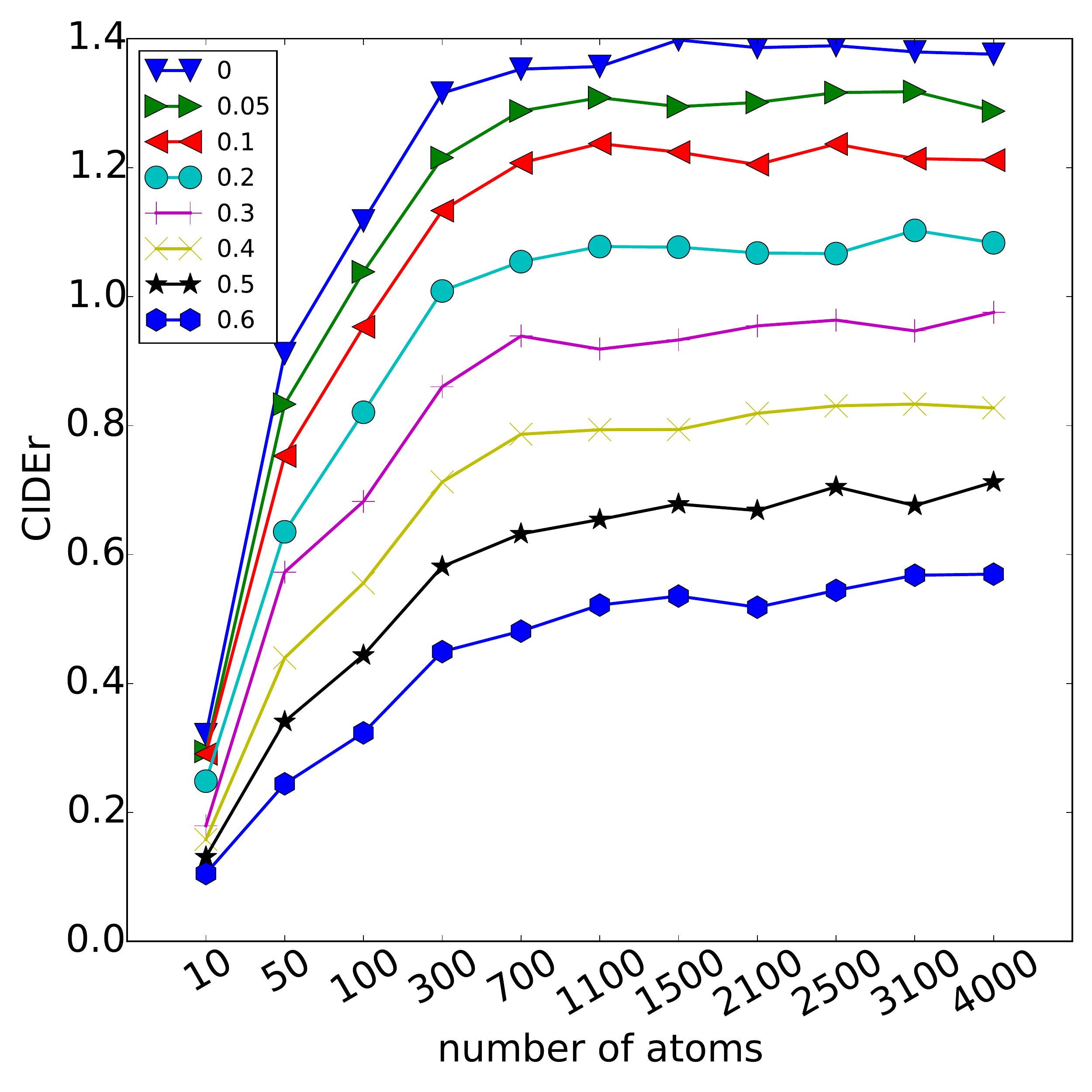}
\end{minipage}\hfill
\begin{minipage}{0.3\textwidth}
\includegraphics[scale=0.15]{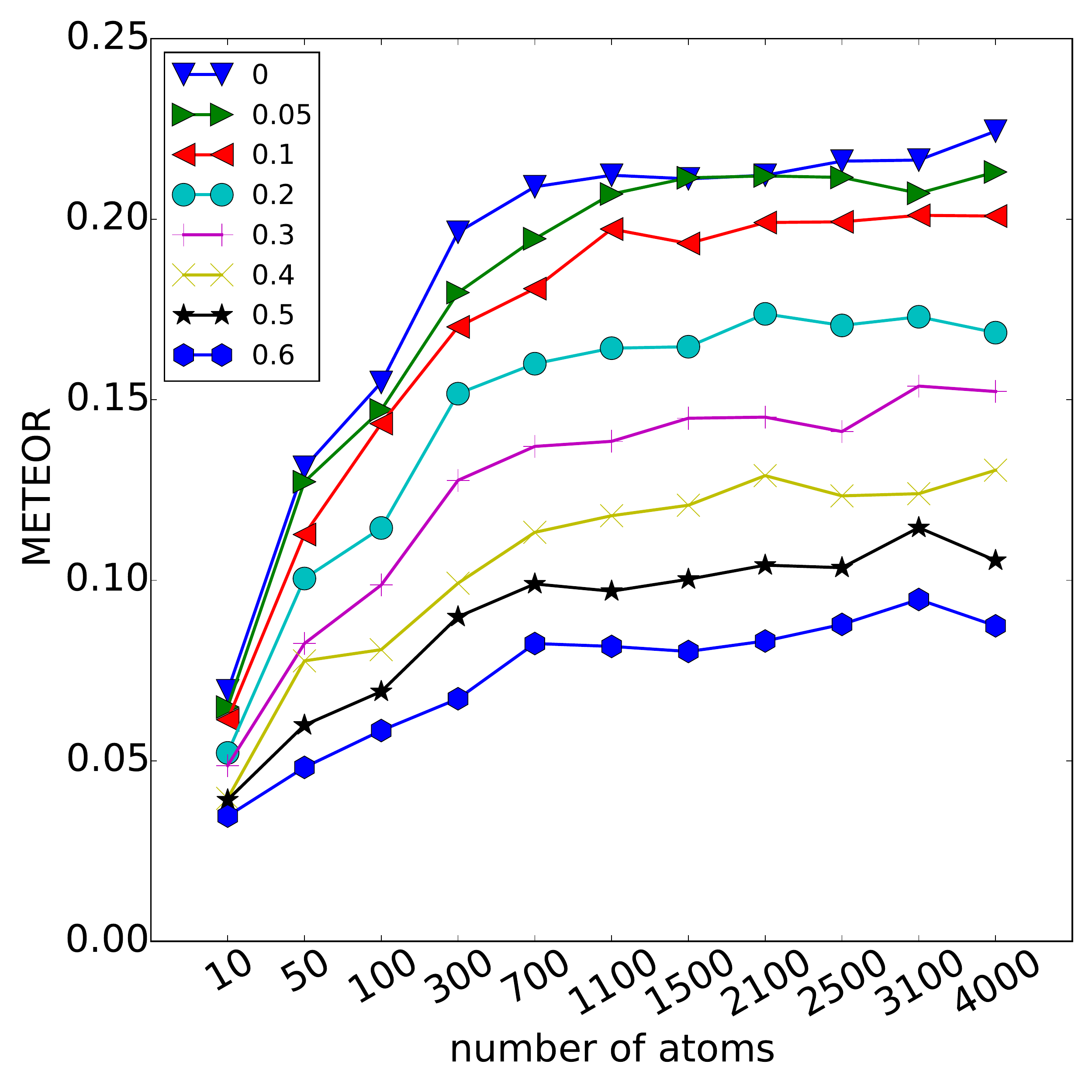}
\end{minipage}\hfill
\begin{minipage}{0.4\textwidth}
  \scriptsize{
  \caption{Learned oracles with different atom precision ($r$ in red)
    and atom quantity (x-axis) on COCO (left) and LSMDC (right).
The number of atoms $\hat{\va}^{(k)}_r$ is varied on x-axis and oracles 
are computed on y-axis on testsets. CIDEr is used for 
COCO and METEOR for LSMDC.
It shows one could increase 
the score by either improving  $P(\mathbf{a}^{(k)}|\mathbf{v})$ with a fixed $k$ or increase $k$. 
It also shows the maximal error bearable for different score.}
  \label{fig_noisy_atoms}}
\end{minipage}
\end{figure}

Although visual atoms dominant the three atom categories shown in Section \ref{sec_human_eval}, 
as they increase in number, more and more non-visual atoms may be 
included, such as ``living'', ``try'', ``find'' and ``free'' which are relatively difficult to be associated 
with a particular part of visual inputs. Excluding non-visual atoms in the 
conditional language model can further tighten the oracle bound as less hints are provided to it. 
The major difficulty lies in the labor of hand-separating visual atoms from non-visual ones as to the our best 
knowledge this is difficult to automate with heuristics.
\paragraph{atom accuracy versus atom quantity}\label{sec_atom_acu_qua}
We have assumed that 
the atoms are given, or in other words, the prediction accuracy of atoms 
is 100\%. 
In reality, one would hardly expect to have a perfect atom classifier. There is 
naturally a trade-off between number of atoms one would like to capture and the 
prediction accuracy of it. Figure \ref{fig_noisy_atoms} quantifies this trade-off 
on COCO and LSMDC. It also indicates the upper limit of performance given different 
level of atom prediction accuracy. In particular, we have replaced $\va^{(k)}$ 
by $\hat{\va}^{(k)}_r$ where $r$ portion of 
$\va^{(k)}$ are randomly selected and replaced by other randomly picked atoms not appearing in 
$\va^{(k)}$. The case of $r=0$ corresponds to those shown in Figure \ref{fig_overall}. 
And the larger the ratio $r$, the worse the assumed atom prediction is. The value of $r$ 
is shown in the legend of Figure \ref{fig_noisy_atoms}. According to the figure, 
in order to improve the caption generation score, 
one would have two options, either by keeping the number of atoms fixed while improving the 
atom prediction accuracy or by keeping the accuracy while increasing the number of included atoms.
As state-of-art visual model already model around 1000 atoms, we hyphotesize that we could gain more in
improving the atoms accuracy rather than increase the number of atom detected by those models.
\paragraph{intrinsic difficulties of particular datasets}
Figure \ref{fig_overall} also reveals the intrinsic properties of each dataset. In general,
the bounds on YouTube2Text are much higher than COCO, with LSMDC the lowest. 
For instance, from the first column of the figure, taking 10 atoms 
respectively, BLUE-4 is around 0.15 for COCO, 0.30 for YouTube2Text and less than 0.05 
for LSMDC. With little visual information to condition upon, 
a strong language model is required, which makes a dramatic difference across 
three datasets. 
Therefore the oracle, when compared across 
different datasets, offer an objective measure of difficulties of using them 
in the captioning task. 


\section{Discussion}
This work formulates oracle performance for visual captioning.  
The oracle is constructed with the assumption of decomposing visual captioning 
into two consecutive steps.
We have assumed the perfection of the first step where visual atoms are recognized, 
followed by the second step where language models conditioned on visual atoms are trained to 
maximize the probability of given captions.
Such an empirical construction requires only automatic atom parsing
and the training of conditional language models, without extra labeling or costly human evaluation.

Such an oracle enables us to gain insight on several 
important factors accounting for both success and failure of the current state-of-the-art models. 
It further reveals model independent properties on different datasets.
We furthermore relax the assumption of prefect atom prediction. This 
sheds light on a trade-off between atom accuracy and atom coverage, providing guidance to future 
research in this direction. Importantly, our experimental results suggest that
more efforts are required in step one where visual inputs are converted into visual concepts (atoms).

Despite its effectiveness shown in the experiments, the empirical oracle is constructed with
the simplest atom extraction procedure and model parameterization in mind, which makes such a
construction in a sense a ``conservative'' oracle.


 

\section*{Acknowledgments}

The authors would like to acknowledge the support of the following agencies for
research funding and computing support: IBM T.J. Watson Research, NSERC, Calcul Qu\'{e}bec, Compute Canada,
the Canada Research Chairs and CIFAR. We
would also like to thank the developers of Theano \citep{2016arXiv160502688short}
, for developing such a
powerful tool for scientific computing.

\bibliography{paper}
\end{document}